\documentclass[conference]{IEEEtran}
\IEEEoverridecommandlockouts
\usepackage{cite}
\usepackage{amsmath,amssymb,amsfonts}
\usepackage{algorithmic}
\usepackage{graphicx}
\usepackage{textcomp}
\usepackage{xcolor}
\usepackage{subcaption}
\usepackage{graphics}
\usepackage{bm}
\usepackage{svg}
\usepackage{multirow}
\usepackage{url}
\usepackage{pifont}
\usepackage{scalerel}
\usepackage{bbm}
\usepackage[ruled,vlined,linesnumbered,algo2e,noend]{algorithm2e}
\usepackage{adjustbox}
\usepackage{booktabs}
\usepackage{hyperref}

\newcommand*{\affaddr}[1]{#1}
\newcommand*{\affmark}[1][*]{\textsuperscript{#1}}
\newcommand*{\email}[1]{\texttt{#1}}

\DeclareMathOperator{\transformer}{Transformer\_Encoder}
\DeclareMathOperator{\embedding}{Embedding}

\graphicspath{{fig/}}

\def\BibTeX{{\rm B\kern-.05em{\sc i\kern-.025em b}\kern-.08em
    T\kern-.1667em\lower.7ex\hbox{E}\kern-.125emX}}
\begin{document}
\title{Heterogeneous Treatment Effect Estimation with Subpopulation Identification for Personalized Medicine in Opioid Use Disorder}

\author{
Seungyeon Lee\affmark[1,2], Ruoqi Liu\affmark[1,2], Wenyu Song\affmark[3], Ping Zhang\affmark[1,2]\\
\affaddr{\affmark[1]Department of Computer Science and Engineering, The Ohio State University, USA}\\
\affaddr{\affmark[2]Department of Biomedical Informatics, The Ohio State University, USA}\\
\email{\{lee.10029,liu.7324,zhang.10631\}@osu.edu}\\
\affaddr{\affmark[3]Department of Medicine, Brigham and Women’s Hospital, Harvard Medical School, USA}\\
\email{\
wsong@bwh.harvard.edu}
}

\newcommand{\zhangcomment}[1]{\textcolor{red}{#1}}
\newcommand{\songcomment}[1]{\textcolor{blue}{#1}}
\newcommand{\liucomment}[1]{\textcolor{green}{#1}}
\newcommand{\leecomment}[1]{\textcolor{purple}{#1}}

\maketitle

\begin{abstract}
Deep learning models have demonstrated promising results in estimating treatment effects (TEE). However, most of them overlook the variations in treatment outcomes among subgroups with distinct characteristics. This limitation hinders their ability to provide accurate estimations and treatment recommendations for specific subgroups. In this study, we introduce a novel neural network-based framework, named SubgroupTE, which incorporates subgroup identification and treatment effect estimation. SubgroupTE identifies diverse subgroups and simultaneously estimates treatment effects for each subgroup, improving the treatment effect estimation by considering the heterogeneity of treatment responses. Comparative experiments on synthetic data show that SubgroupTE outperforms existing models in treatment effect estimation. Furthermore, experiments on a real-world dataset related to opioid use disorder (OUD) demonstrate the potential of our approach to enhance personalized treatment recommendations for OUD patients.
\end{abstract}


\section{Introduction}

Opioid Use Disorder (OUD) is a significant healthcare and economic burden. While there are multiple drugs approved by the Food and Drug Administration (FDA) for the treatment of OUD, including Methadone, Buprenorphine, and Naltrexone, they are either restricted in usage or less effective.  Furthermore, there is a phenomenon where individuals have different treatment responses to the same treatment. Therefore, evaluating approved drugs for OUD and further identifying patient characteristics that may increase the risk of opioid-related adverse events, compared to the general population, is critical to enhance the effectiveness of OUD treatment. Specifically, identifying subgroups of patients with diminished or enhanced treatment responses can provide valuable insights for experts to guide prescribing decisions. 

Numerous deep learning models have shown promising performance in treatment effect estimation (TEE) \cite{curth2021inductive,schwab2020learning, shalit2017estimating,shi2019adapting,nie2021vcnet,zhang2022exploring}. However, most of them ignore heterogeneous subgroups with diverse treatment effects, which reflect the diverse characteristics of populations in real-world settings. This limitation hinders more accurate estimations of treatment effects and treatment recommendations for specific subgroups. Heterogeneous treatment effect (HTE) analysis, also known as subgroup analysis, addresses the heterogeneity of treatment outcomes by identifying subgroups with similar covariates and/or treatment responses. However, the existing subgrouping methods \cite{yang2022tree,loh2016identification, foster2011subgroup, lee2020causal, lee2020robust, argaw2022identifying, nagpal2020interpretable} have limitations; many of these approaches i) rely on traditional machine learning models, which may struggle when dealing with high-dimensional data, and ii) require a one-time pre-estimation of treatment effects to identify heterogeneous subgroups, which can result in suboptimal performance if the pre-estimation is inaccurate.

To tackle these challenges, we propose a novel neural network-based framework, named SubgroupTE. This framework seamlessly integrates subgroup identification and treatment effect estimation to identify heterogeneous subgroups with diverse treatment responses, rather than treating the entire population as a homogenous group. SubgroupTE incorporates subgroup information directly into the estimation process, leading to more accurate treatment effectiveness estimates and enabling personalized treatment recommendations for specific subgroups. Furthermore, SubgroupTE leverages an expectation–maximization (EM)-based training process that optimizes both the subgroup identification and treatment effect estimation networks, ultimately enhancing estimation accuracy and subgroup identification. This approach effectively addresses the limitations associated with the one-time pre-estimation step in the existing subgrouping methods, resulting in improved overall model performance. The SubgroupTE framework is composed of three key components: (i) a Feature Representation Network that learns latent representations from input data, (ii) a Subgrouping Model that identifies patient subgroups with enhanced or diminished treatment effects, and (iii) a Subgroup-informed prediction network that predicts treatment effects based on subgroup information.

In summary, our contributions are as follows:

\begin{itemize}
    \item We develop a model that incorporates subgroup identification and treatment effect estimation. This model enhances the accuracy of treatment effect estimation by considering subgroup-specific information into the estimation process.
    \item We design an EM-based training process that iteratively train both the subgrouping and treatment effect estimation networks, ultimately leading to improved estimation accuracy and more precise subgroup identification.
    \item We demonstrate the capability of our approach to improve the personalized treatment selection for OUD patients in real-world problems.     

\end{itemize}

\section{Related works}\label{sec2}
\textbf{Treatment effect estimation.}
Numerous endeavors have been made to harness the potential of neural networks for treatment effect estimation. In prior research \cite{shi2019adapting,curth2021inductive,schwab2020learning, shalit2017estimating}, a key emphasis has been on distinguishing the treatment variable from other covariates to ensure that treatment information is preserved within the high-dimensional latent representation. However, many of these approaches tend to overlook the variability in treatment effects across distinct subgroups. This limitation can impede their ability to make accurate predictions of treatment effects and offer personalized treatment recommendations for specific groups. In contrast, our proposed method tackles this issue by simultaneously addressing subgroup identification and treatment effect estimation.

\textbf{Subgroup analysis for treatment effects.} 
Subgroup analysis for causal inference focuses on identifying subgroups whose subjects have similar characteristics and/or responses to treatment. The conventional approach identifies subgroups by optimizing heterogeneity/homogeneity of treatment effects across/within these subgroups using one-time pre-estimation of potential outcomes \cite{yang2022tree,loh2016identification, foster2011subgroup, lee2020causal, lee2020robust, argaw2022identifying}. However, this approach heavily depends on the quality of the one-time pre-estimation step, which can be problematic in certain scenarios and result in suboptimal outcomes if the initial estimation is inaccurate. Our proposed model addresses the challenges associated with the quality of the pre-estimation step by employing an EM-based training process that iteratively trains both treatment effect estimation and subgroup identification models.

\begin{figure*}[t]
\centering
\includegraphics[width=0.9\linewidth]{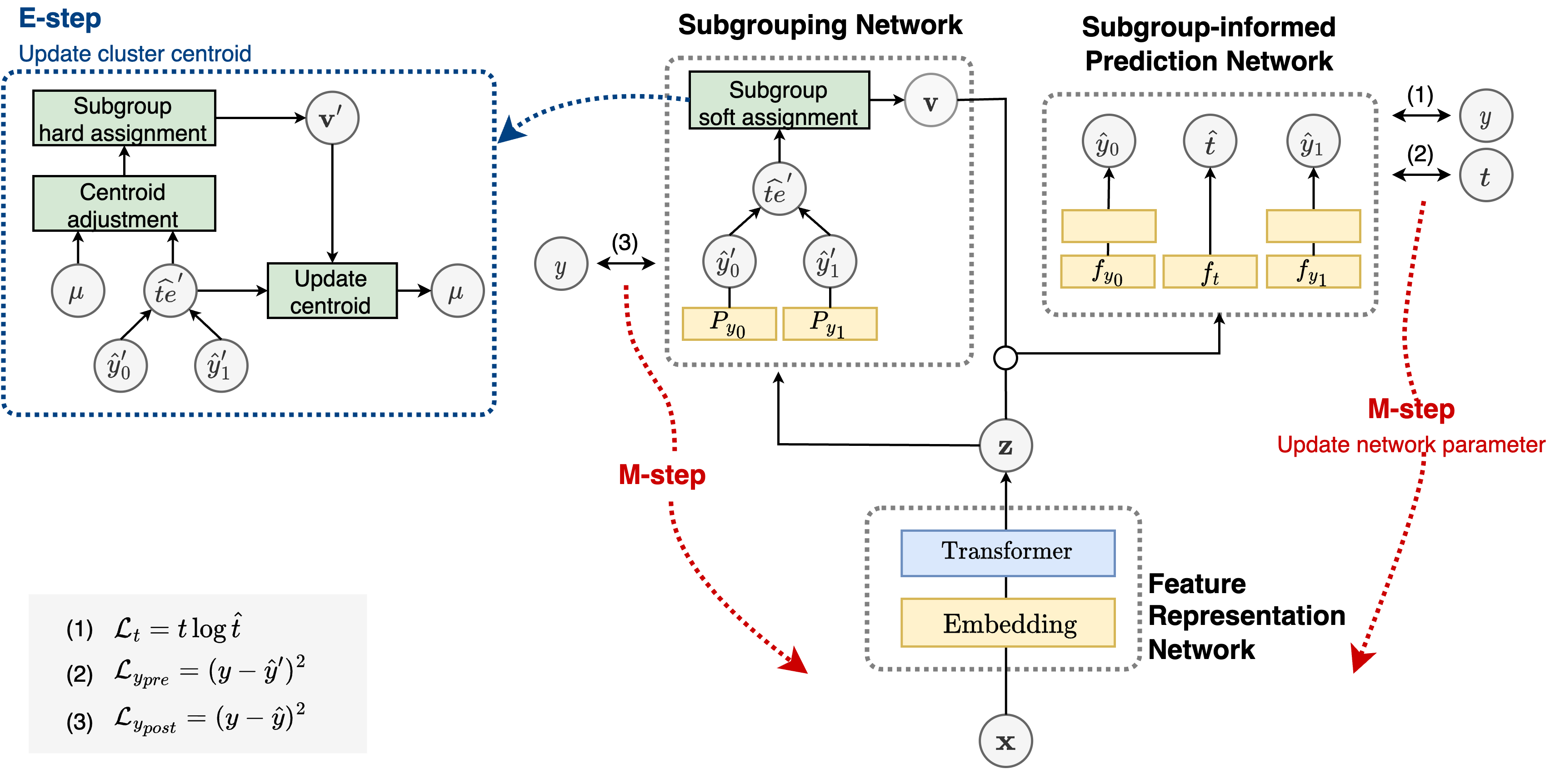}
\caption{Architecture of SubgroupTE.}
\label{fig1}
\end{figure*}

\section{Methodology}\label{sec3}

\subsection{Problem statement and assumptions}
In our scenario, we have a dataset of $N$ observed samples denoted as $D \equiv \{(\mathbf{x}_i,t_i,y_i)\}^N_{i=1}$, where $\mathbf{x}_i \in \mathbb{R}^p$ represents pre-treatment covariates, $t_i$ is the treatment assignment, which has binary values \{0,1\} with a binary treatment setting, and $y_i$ denotes the potential outcome, which is the response of the $i$-th sample to the treatment $t_i$. The propensity score is defined as the conditional probability of receiving treatment given the observed covariates, expressed as $p(t=1|\mathbf{x})$. The primary objective of the predictive model is to predict these potential outcomes. Within the framework of potential outcomes introduced by \cite{rubin1974estimating}, we define treatment effect as the difference between potential outcomes with and without treatment, expressed as $\mathbb{E}[Y(1) - Y(0) |\mathbf{x}]$. To estimate the treatment effect, we rely on the assumption of unconfoundedness, which consists of three main components: i) Conditional Independence Assumption, ii) Common Support Assumption, and iii) Stable Unit Treatment Value Assumption. These assumptions collectively provide the necessary conditions for obtaining unbiased and consistent estimations of causal effects.

\subsection{Proposed model}

SubgroupTE, our proposed framework, consists of three main networks: the feature representation network, which maps the input data into latent representations for treatment effect estimation; the subgrouping network, which pre-estimates potential outcomes and assigns subgroup probabilities to each data point; and the subgroup-informed prediction network, which performs the final estimation of potential outcomes while incorporating subgroup information into the estimation process. SubgroupTE utilizes an Expectation-Maximization (EM)-based training process that iteratively trains network parameters and cluster centroids for subgrouping. An overview of the SubgroupTE framework is provided in Figure \ref{fig1}.

\subsubsection{\textbf{Feature representation network}} To construct the feature representation network $Q_{\phi}$, we utilize an embedding layer and an encoder network of the Transformer. Given a data sample $\textbf{x}_i$, it is input into the embedding layer, and subsequently, its output is fed into $\transformer$ to extract the latent representations.

\begin{align}
    \textbf{z}_i &= Q_{\phi}(\textbf{x}_i) \\
    &= \transformer\left(\embedding(\textbf{x}_i)\right) \nonumber
\end{align}\label{eq1}

\subsubsection{\textbf{Subgrouping model}} The subgrouping model aims to identify patient subgroups whose subjects have similar treatment effects. The model initially estimates control outcome $\hat{y}'_{0}$ and treated outcome $\hat{y}'_{1}$, which are pre-subgrouping estimations, by two separate one-layer feedforward networks, $P_{y_0}$ and $P_{y_1}$. Subsequently, the pre-subgrouping treatment effect $\widehat{te}'=\hat{y}'_{1}-\hat{y}'_{0}$ is computed. The subgrouping model assigns a subgroup probability vector to each data sample, representing the likelihood of that sample belonging to each subgroup.

The subgroup probability vector, denoted as $\mathbf{v}\in\mathbb{R}^K$ where $K$ represents the number of subgroups, is obtained based on the Euclidean distance between the pre-subgrouping treatment effect and subgroup centroids. For each data sample $i$, the distance $d_i^k$ between $\widehat{te}_i$ and the centroid $\mu_k$ of subgroup $k$ is computed as:

\begin{equation*}
    \widehat{te}'_i = P_{y_1}(\textbf{z}_i) - P_{y_0}(\textbf{z}_i)
\end{equation*}

\begin{equation}\label{eq2}
    d_i^k= \left \|\widehat{te}'_i-\mu_k \right \|_2
\end{equation}

The subgroup probability $v_i^k$ indicating the likelihood of data sample $i$ belonging to subgroup $k$ is computed by taking the inverse of the distance $d_i^k$:

\begin{equation}\label{eq3}
    v_{i,k}=\frac{\exp^{-d_i^k}}{\sum_{j=1}^K \exp^{-d_i^j}}
\end{equation}

When the treatment effect $\widehat{te}'_i$ is closer to the centroid $\mu_k$, the corresponding subgroup probability $v_{i,k}$ will have a higher value. The subgroup probability vector $\mathbf{v}$ is then used as additional input features for the subgroup-informed prediction models.

\subsubsection{\textbf{Subgroup-informed prediction network}}
The subgroup-informed prediction network enhances the accuracy of treatment effect estimation by considering the subgroup information. To incorporate the subgroup information into the estimation process, we concatenate two vectors, $\textbf{z}_{i}$ and $\textbf{v}_{i}$, and then we use it as input for the subgroup-informed prediction network. To preserve the treatment information in the high-dimensional latent representation, we construct the subgroup-informed prediction network $f$ with three distinct feedforward networks: $f_{y_0}$, $f_{y_1}$, and $f_{t}$. Each of these networks is responsible for predicting a specific outcome: $f_{y_0}$ for the control outcome $\hat{y}_{0}$, $f{y_1}$ for the treated outcome $\hat{y}_{1}$, and $f_{t}$ for the treatment assignment $\hat{t}$.

\subsubsection{\textbf{Optimization}}
We implement the EM-based training process that iteratively updates both the cluster centroids and the network parameters to train the proposed SubgroupTE.

\textbf{E-step} updates the cluster centroids based on the k-means algorithm to effectively group patients into $K$ subgroups by maximizing the homogeneity of the estimated treatment effects. During this step, all network parameters are fixed. However, as network parameters are updated, the distribution of the feature space—the pre-subgrouping treatment effect estimation—may shift. This shift can result in disparities between the distributions of previously updated centroids and data samples in the new feature space, possibly leading to some clusters containing no data samples. This, in turn, reduces the number of clusters. To tackle this challenge, we utilize a two-step approach. First, we align the distribution of the existing centroids to the new feature space before assigning data points to clusters. This alignment is achieved by computing the Kernel Density Estimation (KDE) between the distributions of the existing centroids and the data samples in the new feature space, and then updating the centroids accordingly using KDE. Specifically, for each centroid, we calculate the KDE between the centroid and data samples. This KDE quantifies the shift in distribution caused by changes in network parameters. Based on the KDE, we adjust the position of each centroid. Secondly, we assign data points to the updated centroids. This approach ensures that the clusters adapt to the new feature space and the clustering remains effective even as the network parameters evolve during training.

The centroid adjustment process is as follows:
\begin{equation*}
\textrm{Kernel}(\widehat{te}'_i,\mu_k) = \frac{e^{-\frac{1}{2} ((\widehat{te}'_i-\mu_{k})/\cdot h)^2}}{\sum_i e^{-\frac{1}{2} ((\widehat{te}'_i-\mu_{k})/\cdot h)^2}}
\end{equation*}

\begin{equation*}
\textrm{Diff}(\widehat{te}', \mu_k) = \sum_i \\ 
\textrm{Kernel}(\widehat{te}'_i,\mu_k)\cdot (\widehat{te}'_i-\mu_k)
\end{equation*}

\begin{equation}\label{eq4}
\mu^*_k = \mu_k + \textrm{Diff}(\widehat{te}', \mu_k)
\end{equation}

The process involves calculating a weight for each data sample that reflects its proximity to the cluster centroids. Data points closer to the cluster centroid receive higher weights, whereas those farther away receive lower weights. To adjust the centroids, we multiply the difference between the data points and each centroid by their weights and sum them across all data points, which is denoted as $\textrm{Diff}(\cdot)$. Consequently, we adjust the current position of cluster centroid to move toward high-density regions in the feature space.

The centroid is then updated using the following equations for a given mini-batch. The hard assignment vector $v'_i$ is first computed as:

\begin{equation}\label{eq5}
    v'_{i,j}= \left\{\begin{matrix}
    1, & j = \textrm{argmin}_{k=\{1,...,K\}} {\left \| \widehat{te}'_i-\mu^*_k \right \|_2}, \\ 
    0, & otherwise.
    \end{matrix}\right.
\end{equation}

\begin{align}\label{eq6}
\mu_{k}= \left\{\begin{matrix}
\frac{1}{|B_k|}\sum_{i\in B_k}\widehat{te}'_i,&|B_k|>0 \\ 
\mu^*_k, & otherwise.
\end{matrix}\right.
\end{align}
where $B_k$ indicates all the samples assigned to $k$-th cluster such that $B_k = \{i \mid \forall i, v'_{i,k}=1\}$.

The \textbf{M-step} updates the network parameters while keeping the cluster centroids fixed. The subgroup probability vector $\mathbf{v}$ is assigned to each data sample using Eq. (\ref{eq3}). Subsequently, the network parameters are updated based on the predictions generated by the subgrouping and subgroup-informed prediction networks. A loss function is defined as follows:

\begin{align}
\label{eq7}    \mathcal{L} = \alpha \cdot \sum_{i=1}^{|D|}t_i\log{\hat{t}_i} +\beta \cdot \sum_{i=1}^{|D|}(y_i-\hat{y}'_i)^{2} + \gamma \cdot \sum_{i=1}^{|D|}(y_i-\hat{y}_i)^{2} \nonumber
\end{align}
where, $\hat{y}'_{i}$ and $\hat{y}_{i}$ represent the pre- and post-subgrouping estimations of the factual outcomes for the $i$-th sample. These estimations are obtained from the subgrouping and subgroup-informed prediction networks, respectively. $\alpha$, $\beta$, and $\gamma$ are hyper-parameters.

\section{Experiments}\label{sec4}

\subsection{Datasets}
Our experiments are conducted on two datasets: synthetic and real-world datasets. The \textbf{synthetic dataset}, which includes both treated and control outcomes, is utilized to assess the accuracy of treatment effect estimation. We generate the dataset, following previous studies \cite{lee2020robust, argaw2022identifying}, which is inspired by the initial clinical trial results of remdesivir for COVID-19 treatment \cite{wang2020remdesivir}. It comprises 10 covariates and outcomes simulated using the 'Response Surface B' model proposed in \cite{hill2011bayesian}. We randomly generate a total of 1,000 samples, consisting of 500 treated and 500 control patients.

\subsection{Experimental setup}

\begin{table}[]
\centering
\caption{Prediction results on the synthetic dataset. The average score and standard deviation under 30 trials are reported.}\label{tb3}
\begin{tabular}{l|cc}\toprule
\multicolumn{1}{c|}{Dataset}     & \multicolumn{2}{c}{Synthetic}                                                        \\
\multicolumn{1}{c|}{Model}       & PEHE                                        & $\epsilon$ATE  \\ \midrule
RF         & 0.086 $\pm$ 0.000                           & 0.039 $\pm$ 0.000                                                  \\
SVR        & 0.103 $\pm$ 0.000                           & 0.029 $\pm$ 0.000                                     \\ \midrule
SubgroupTE & \textbf{0.024 $\pm$ 0.002} & \textbf{0.014 $\pm$ 0.009} \\ \bottomrule
\end{tabular}
\end{table}

\begin{figure}[!t]
\centering
\includegraphics[width=0.9\linewidth]{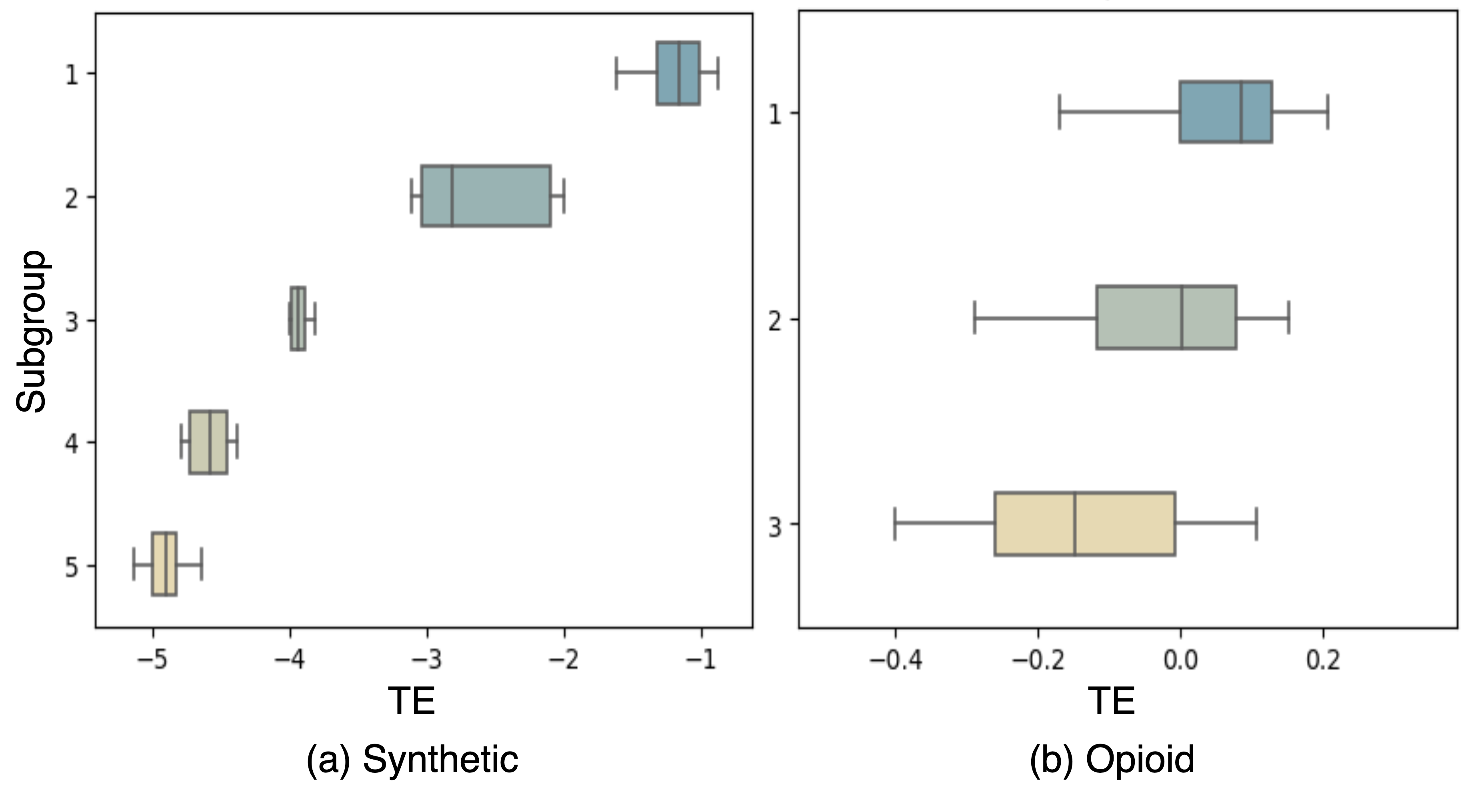}
\caption{Visualization of the treatment effect distribution for the identified subgroups on (a) synthetic and (b) opioid datasets. Each box signifies the interquartile range, spanning from the 25th to the 75th percentiles of the treatment effect distribution. The whiskers cover the range between the 5th and 95th percentiles.} \label{fig3}
\end{figure}


\textbf{Implementation details.} In our comparison study, we evaluate the performance of SubgroupTE compared to two machine learning methods, random forest (RF) and support vector machine (SVR). The SubgroupTE is implemented using PyTorch. For the optimization, we employ the SGD algorithm with a mini-batch size of 64 and a learning rate of 0.001. The number of hidden nodes is set from \{50, 100, 200, 300\}. We also explore a range of coefficients ($\alpha$, $\beta$, $\gamma$) within the interval [0, 1] and the number of subgroups within the range of [1, 10]. We split the dataset with a ratio of 6 : 2 : 2 for training, validation, and test datasets, respectively. 

\textbf{Evaluation metric.} In our evaluation, we utilize two key metrics to assess the performance of our model. Heterogeneous effects (PEHE) measures the accuracy of estimating treatment effects at the individual level. It is defined as $\textrm{PEHE} = \frac{1}{N}\sum_{i=1}^N (f_{y_1}(\mathbf{x}_i)-f_{y_0}(\mathbf{x}_i)-\mathbb{E}[Y_1-Y_0|\mathbf{x}_i])^2$. $\epsilon$ATE assesses the overall treatment effect at the population level and is defined as $\textrm{$\epsilon$ATE} = \left |\mathbb{E}[f_{y_1}(\mathbf{x})-f_{y_0}(\mathbf{x})]-\mathbb{E}[Y_1-Y_0] \right |$.

\subsection{Results on synthetic data}

Table~\ref{tb3} presents the prediction results on the synthetic dataset. SubgroupTE outperforms other baseline models in terms of treatment effect estimation. Specifically, it achieves a PEHE score of 0.024, which represents a 76.7\% reduction compared to the second-best model. These improvements highlight the advantages of SubgroupTE. By incorporating subgroup information into the estimation process, our model achieves a more precise estimation of treatment effects.

To assess whether the identified subgroups are appropriate, we visualize boxplots of the treatment effect distribution for these subgroups in Fig. \ref{fig3} (a). SubgroupTE effectively identifies subgroups, as evidenced by the significant difference in average treatment effects among subgroups and the clear separation of their distributions. These results provide strong evidence supporting the effectiveness of SubgroupTE in accurately identifying subgroups with diverse treatment effects.

\subsection{Real-world study}
\textbf{Problem statement and dataset.} OUD represents a substantial challenge in both healthcare and economic terms. Despite various approved drugs for OUD treatment, such as Methadone, Buprenorphine, and Naltrexone, they are either restricted in usage or less effective. While Naltrexone can be prescribed by any licensed medical practitioner, Methadone and Buprenorphine have regulatory constraints\cite{sharma2017update}; Methadone is available exclusively through regulated Opioid Treatment Programs (OTPs), and Buprenorphine can be prescribed by physicians who have completed specific training or possess addiction board certification and have obtained a federal waiver. In addition, there is widely acknowledged empirical evidence supporting the effectiveness of MOUD, specifically Methadone or Buprenorphine\cite{heidbreder2023history}. In light of this, we compare these two well-established drugs, Methadone and Buprenorphine, with Naltrexone, a newly approved treatment, to assess its relative efficacy and suitability for OUD treatment. We sample around 600K distinct OUD patients from the MarketScan Commercial Claims and Encounters (CCAE) \cite{market} from 2012 to 2017. OUD patients are identified using opioid-related emergency department (ED) visits. The MarketScan claims data provide patients' medical histories, including diagnoses, procedures, prescriptions, and demographic characteristics.

\begin{figure}[!t]
\centering
\includegraphics[width=1\linewidth]{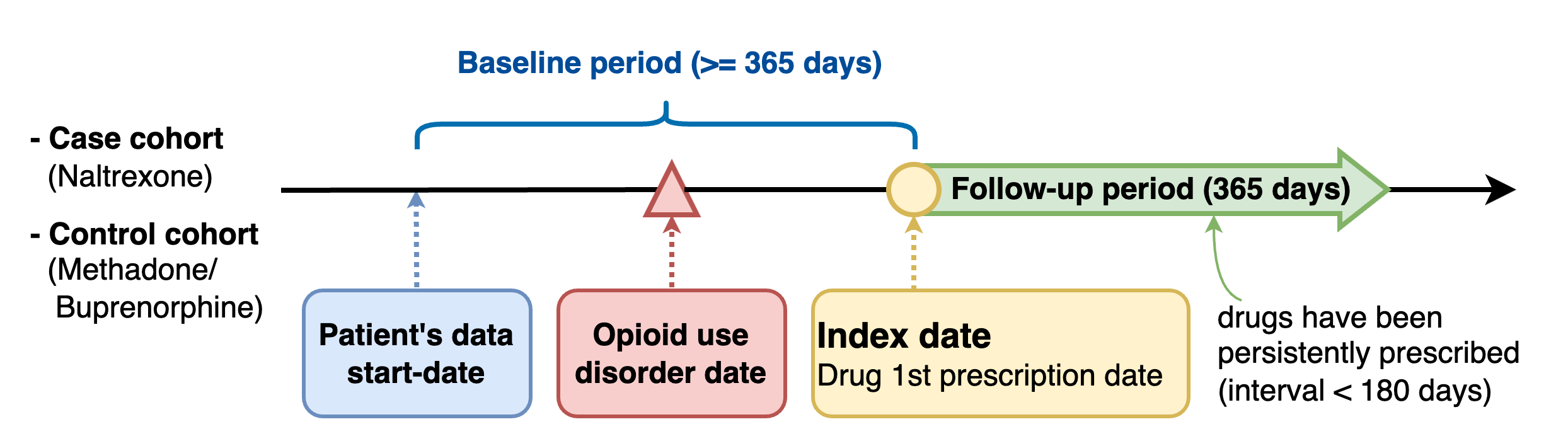}

\caption{Illustration of the overall design. The index date refers to the first prescription date of the drug. Baseline and follow-up periods include all the dates before and after the index date, respectively.}\label{fig5}
\end{figure}

\textbf{Study design.} This study aims to evaluate the effect of Naltrexone and identify subgroups at high/low risk of OUD-related adverse outcomes. To define these outcomes, we consulted domain experts to define clinically relevant events associated with OUD \cite{urman2021burden, sun2022evaluation, zhang2022examining, powell2023variation}, such as opioid overdose, Opioid-Related Adverse Drug Events (ORADEs), and hospitalization. Opioid overdose and ORADEs are identified based on diagnosis codes. Hospitalization refers to inpatient visits and is determined by whether or not the patient is hospitalized after taking the drugs. We label all patients who have these adverse events as positive, otherwise negative. We use the occurrence rates of adverse events to estimate treatment effects. For evaluating the effect of Naltrexone, we include patients in a case cohort if prescribed with Naltrexone, and in a control cohort if prescribed with Methadone or Buprenorphine. The statistics are described in Table \ref{tb6} and the detailed cohort selection criteria are shown in Fig. \ref{fig5}. We use diagnosis codes and medication information to construct covariates in our cohort. The diagnosis codes are defined by International Classification of Diseases (ICD) 9/10 codes. We map the ICD codes to Clinical Classifications Software (CCS), including a total of 286 codes. For the medications, we match national drug codes (NDCs) to observational medical outcomes partnership (OMOP) ingredient concept IDs, resulting in a total of 1,353 unique drugs in our dataset.

\begin{table}[!t]
\caption{Statistics on OUD dataset}\label{tb6}
\centering
\begin{tabular}{l|ccc}
\toprule
                & $Y=0$ & $Y=1$ & Total \\ \midrule
Case ($T=1$)    & 403   & 353   & 756   \\
Control ($T=0$) & 1,430 & 707   & 2,137 \\
Total           & 1,833 & 1,060 & 2,893 \\ \bottomrule
\end{tabular}

\end{table}

\textbf{Results.} We visualize the distribution of treatment effects for the identified subgroups in Fig. \ref{fig3} (b). The third subgroup, characterized by a negative average treatment effect and the smallest values, shows the most enhanced effect among all subgroups. This suggests that Naltroxene has a more favorable effect compared to Methadone and Buprenorphine for this subgroup. The second subgroup has an average treatment effect close to zero, indicating that Naltroxene has minimal impact on the outcome. On the other hand, the first subgroup exhibits a positive averaged treatment effect, indicating a diminished effect for Naltroxene. Therefore, Naltroxene would be recommended for the third subgroup.

\begin{figure}[!th]
\includegraphics[width=0.9\linewidth]{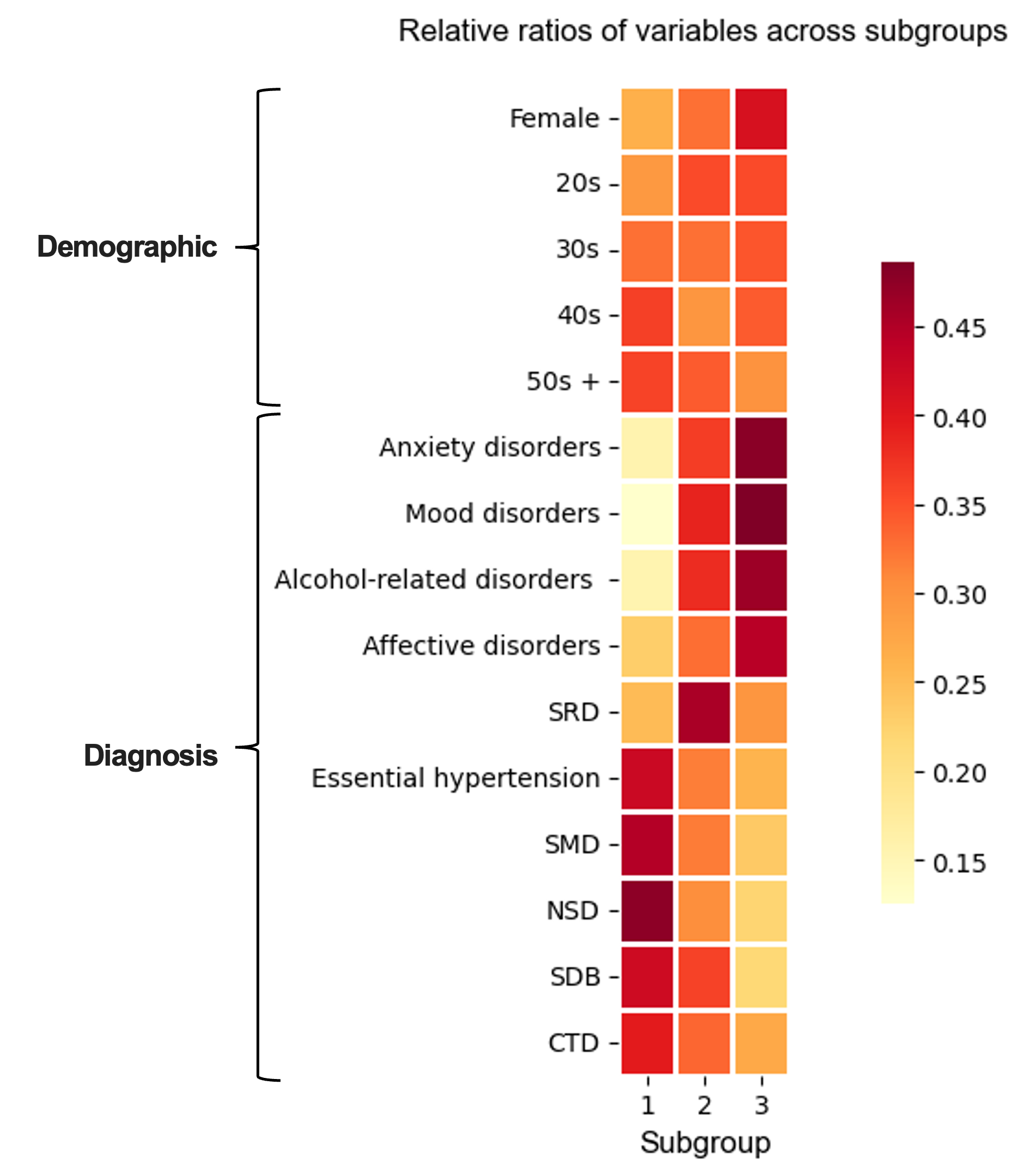}
\caption{The heatmap of the relative ratios for the variables related to demographics and diagnosis codes across the three subgroups. The relative ratio is computed as $\pi_{k,i}/\sum_{k=1}^{K}\pi_{k,i}$, where $\pi_{k,i}$ is the ratio of the $i$-th variable in the $k$-th subgroup. SMD: Substance-related mental disorders; NSD: Other nervous system disorders; SDB: Spondylosis, intervertebral disc disorders, or other back problems; CTB: Other connective tissue disease.
}\label{fig7}
\end{figure}

To further compare the identified subgroups, we analyze the variables related to demographics and diagnosis codes for all subgroups. Fig. \ref{fig7} presents a heatmap showing the relative ratios of these variables across the three subgroups. To calculate the relative ratios, we compute the ratio of each variable for each subgroup and then scale the ratios across all subgroups. For the diagnosis codes, we consider only those recorded during the baseline period, which is used to estimate the potential outcomes. We identify the top 15 most frequent codes and select the 10 most differentiated codes across subgroups. In Fig. \ref{fig7}, we observe that as the treatment effect improves, the ratios of the diagnosis codes generally exhibit a sequential increase or decrease. In addition, the third subgroup, which has the most enhanced treatment effect, has a higher proportion of females and a larger distribution of younger patients. These findings demonstrate that the identified subgroups are clinically distinct and the proposed model effectively identifies subgroups based on patients' medical history and demographic information.

\section{Conclusion}
In this work, we address the critical problem of estimating treatment effects, particularly when leveraging real-world clinical datasets. We introduce a novel framework to account for the heterogeneity of responses within the population. This framework seamlessly integrates two critical components: subgroup identification and treatment effect estimation. By doing so, it effectively navigates and addresses the inherent heterogeneity observed within patient populations – a heterogeneity that often complicates treatment optimization and personalization. The results derived through the empirical experiments provide robust and compelling evidence supporting the effectiveness of our approach. They demonstrate that our method not only identifies subgroups within the patient population but also accurately estimates the treatment effects by subgroup. Further augmenting the credibility of our approach is a real-world study. This study serves as a practical demonstration of how our framework can be applied in a clinical setting, showing its utility in enhancing personalized treatment selection and optimization.

\bibliographystyle{IEEEtran}
\bibliography{ref}

\end{document}


\section{Dataset}
The synthetic data is inspired by the initial clinical trial results of remdesivir to COVID-19 \cite{wang2020remdesivir}. The clinical trials show that remdesivir results in faster clinical improvement in patients with shorter time from symptom onset to trial start. The synthetic data consists of 10 covariates used in the trial. All covariates are randomly simulated from Normal distribution with different parameter values. To follow the trial results, the potential outcomes are simulated by adapting the standard non-linear 'Response Surface B' proposed in \cite{hill2011bayesian} for the covariates except for the time. The time covariate is fed into a logistic function to produce the faster clinical improvement times with the shorter time from symptom onset to trial start. The potential outcomes are as follow. We randomly generate the total number of 1,000 samples with 500 case 
 ($T=1$) and control ($T=0$) patients.

\begin{align*}
    &Y_0\sim\mathcal{N}(X_{-0}\mathbf{\beta}+(1+e^{-(x_0+9)})^{-1}+5, 0.1)\\
    &Y_1\sim\mathcal{N}(X_{-0}\mathbf{\beta}+5\cdot(1+e^{-(x_0+9)})^{-1}, 0.1)
\end{align*}

\bibliographystyle{IEEEtran}
\bibliography{ref}